# Automatic Ranking of MT Outputs using Approximations

Pooja Gupta
Apaji Institute
Banasthali University
Rajasthan, India
poojagupta2291@gmail.com

Nisheeth Joshi
Apaji Institute
Banasthali University
Rajasthan, India
nisheeth.joshi@rediffmail.com

Iti Mathur
Apaji Institute
Banasthali University
Rajasthan, India
mathur_iti@rediffmail.com

## ABSTRACT
Since long, research on machine translation has been ongoing. Still, we do not get good translations from MT engines so developed. Manual ranking of these outputs tends to be very time consuming and expensive. Identifying which one is better or worse than the others is a very taxing task. In this paper, we show an approach which can provide automatic ranks to MT outputs (translations) taken from different MT Engines and which is based on N-gram approximations. We provide a solution where no human intervention is required for ranking systems. Further we also show the evaluations of our results which show equivalent results as that of human ranking.

## 1. INTRODUCTION
Ngram approximation is the subtask of Natural Language Processing (NLP), which is the branch of artificial intelligence. Approximation has many applications mainly in machine translation and natural language processing. In this paper we present an unsupervised learning approach for the development of a Ranking System. For this, we have done our study on English-Hindi language pair. We describe the discriminative training approach of machine learning in detail to identify the best MT Engine Output. The main idea behind the use of MT Engine output is to predict the correct translation of a sentence. Assessing the correct machine translation output is very difficult. There are lots of MT engines being developed in the world and there are various measures through which the quality of machine translation can be computed. With the help of Ranking System we can find out the best and accurate translation in minimum time.

The rest of the paper is organized as follows: Section 2 reviews the work that has been done in this area. Section 3 describes our approach. Section 4 describes the evaluation and results of the study. Finally Section 5 concludes the paper.

## 2. LITERATURE SURVEY
A lot of research work has been done and is still going on in machine translation. As we know that Machine Translation (MT) is becoming very popular among end-users. The main idea of estimating the quality of automatic translations for a particular task is called as the confidence estimation [1]. This confidence estimations task has been transformed into quality estimation task where the central idea remains the same. This quality estimation task is a rather recent aspect in research on Machine Translation.

In this area, previous work includes statistical methods on predicting word-level confidence [2] where just by looking at words people tried to analyze the quality of the translation. This method was extended by Specia et al. [3] who applied a regression technique and used SVM based classifiers. Raybaud et al., 2009) [4] further extended this study by estimating correctness using several probabilistic measures. In this direction, Rosti et al. [5] also performed sentence-level selections with generalized linear models that were based on re-ranking of N-best lists merged from many MT systems. Ye et al. [6] have described machine translation evaluation as a ranking problem as it is often done by the humans. The results show that the greater co-relation with human assessment at the sentence level can be achieved if ranking of translations is done. The authors have also used the n-gram match.

Soricut and Narsal [7] used machine learning for ranking the candidate translations; they then selected the highest-ranked translation as the final output. Avramidis [8] showed an approach of ranking the outputs using grammatical features. He used statistical parser to analyze and generate ranks for several MT outputs. Gupta et al. [9] applied a naïve bayes classifier on English-Hindi Machine Translation System and ranked them. They have used the baseline system that was provided for quality estimation task of WMT 2012 workshop to extract the features of English sentences and its translations produced by MT systems. For evaluating the quality of the systems the authors have used some linguistic features. The authors have also compared the results of the automatic evaluation metrics. Moore and Quirk [10] described smoothing method for N-gram language models based on ordinary counts for generation of language models which can be used for quality estimation task. Setiawan and Zhou [11] employed discriminative training of 150 million translation parameters and its applications to pruning. They had used various pruning techniques for estimation of quality and thus ranking the translations.

## 3. OUR APPROACH
We have used a language model for ranking MT outputs. As language models can very easy capture the structure (grammar) of the language. For this they do not rely on any linguistic analysis but instead requires a large corpus onto which they can apply mathematical models. In this study we have used markov assumption and have used markov chains of order 2.

## 3.1  Experimental Setup

For development of our system, we used 35,000 sentences from tourism domain. These were English sentences with their translations provided by a human. We generated the unigrams, bigrams and trigrams on these 35K sentences. The statistics of this study is shown in table 1. Equations 1, 2 and 3 show the generation of uni, bi and trigrams.

$$P(w_n) = \frac{Count(w_n)}{|V|} \qquad (1)$$

$$P(w_{n-1}w_n) = \frac{Count(w_{n-1}w_n)}{Count(w_{n-1})} \qquad (2)$$

$$P(w_{n-2}w_{n-1}w_n) = \frac{Count(w_{n-2}w_{n-1}w_n)}{Count(w_{n-2}w_{n-1})} \qquad (3)$$

**Table 1. Statistics of the Corpus**

| Corpus | Sentences | Trigrams | Bigrams | Unigrams |
|--------|-----------|----------|---------|----------|
| English | 35000 | 47509 | 272886 | 464969 |
| Hindi | 35000 | 53062 | 308706 | 513910 |

We also used GIZA++ to generate English-Hindi parallel lexicons which we then manually checked and corrected. We used the following algorithm to generate the n-grams for our study. We applied this algorithm on both English as well as Hindi sentences separately.

**Input:** Raw sentences
**Output**: Annotated Text (N-grams text)

**LM Algorithm**

Step1.   Input raw sentence file and repeat steps 2 to 4 for each sentence.
Step2.   Split each word of the sentence.
Step3.   Generate trigrams, bigrams and unigrams for the entire sentence.
Step4.   If n-gram is already present than increase the frequency count.
Step5.   If n-gram is unique than it will sort in descending order by their frequencies.
Step6.   Generate Probability of trigrams using equation 3.
Step7.   Generate Probability of bigrams using equation 2.
Step8.   Generate Probability of unigrams using equation 1.
Step9.   Output obtained in file is in our desired n-garm format.

For our study we have used 1300 English sentences and used six MT engines. The list of engines is shown in table 2. Among these E1, E2 and E3 are MT engines freely available on the internet. E4, E5 and E6 are MT engines that we have developed using different MT toolkits. E4 was a MT system which was trained using Moses MT toolkit [12]. This system used syntax based model [13]. We used Collins parser to generate parses of English sentences and used a tree to string model to train the system. E5 was a simple phrase based MT system which also used Moses MT toolkit. E6 was an example based MT system that was developed by Joshi et al. [14] [15]. These three systems used the 35000 English-Hindi parallel corpora to train and tune themselves. We used 80-20 ratio for training and tuning i.e. we used 28000 sentences to train the systems and remaining 7000 sentences to tune the systems.

## 3.2  Methodology

To rank MT outputs of various systems we first generated the trigrams of English sentence as well as its translations which were produced by different MT engines. To rank the translations we applied the following algorithm:

**Input:** English Sentence with MT outputs
**Output:** Ranked MT output list

**Ranking Algorithm**

Step1.   Trigrams from English sentences are generated.
Step2.   These trigrams are matched with English language model and matched ones are retained.
Step3.   Match retained English trigram's lexicons with English-Hindi parallel lexicon list.
Step4.   If a match is found then register corresponding Hindi lexicon.
Step5.   Match Hindi language model with registered Hindi lexicons and sum the probabilities of each match.
Step6.   Perform these steps on all MT outputs.
Step7.   Sort MT outputs in descending order with respect to their cumulative probabilities.

**Table 2. MT Systems**

| Engine No. | Description |
|------------|-------------|
| E1 | Microsoft Bing MT Engine[1] |
| E2 | Google MT Engine[2] |
| E3 | Babylon MT Engine[3] |
| E4 | Moses Syntax Based Model |
| E5 | Moses Phrase Model |
| E6 | Example Based MT Engine |

Figure 1 shows the working of this entire approach. To have a better understanding of the functionality, we have illustrated the entire process through the following example.

**English Sentence:** Jim Corbett National Park is the oldest national park in India and was established in 1936 as Hailey National Park to protect the endangered Bengal tiger.

**E1 Output:** जिम कॉर्बेट पार्क भारत में सबसे पुराना राष्ट्रीय उद्यान है और 1936 में Hailey राष्ट्रीय उद्यान के रूप में लुप्तप्राय बंगाल बाघ की रक्षा के लिए स्थापित किया गया था।

**E2 Output:** जिम कॉर्बेट नेशनल पार्क भारत में सबसे पुराना राष्ट्रीय उद्यान है और लुप्तप्राय बंगाल टाइगर की रक्षा के लिए हेली नेशनल पार्क के रूप में 1936 में स्थापित किया गया था.

**E3 Output:** जिम कार्बेट राष्ट्रीय उद्यान की स्थापना की गई थी और भारत में सबसे पुराने राष्ट्रीय उद्यान में 1936 में राष्ट्रीय पार्क को बचाने के लिए हेली संकटापन्न बंगाल टाइगर है।

**E4 Output:** जिम कोर्बेट नाशनल पार्क भारत में सबसे पुराना राष्ट्रीय पार्क है और क्या हेल्ली नाशनल पार्क

---

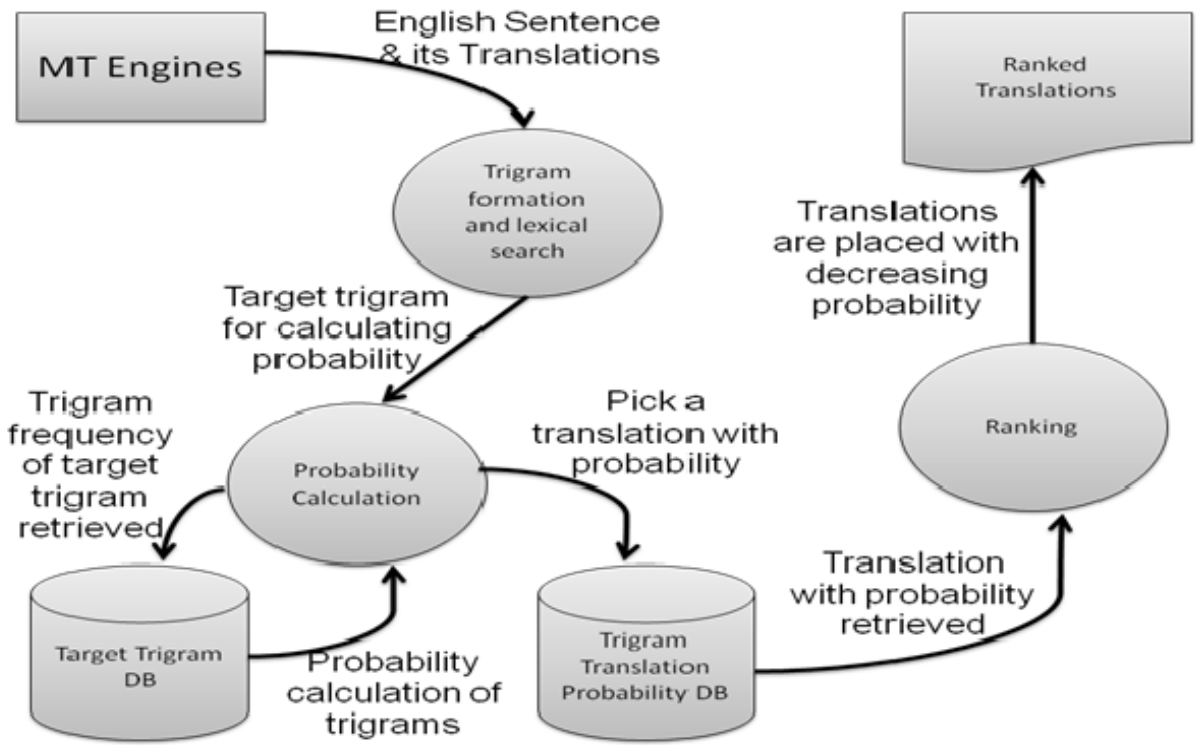

**Figure 1. Ranking System**

की तरह 1936 में स्थापित किया हुआ संकटापन्न बंगाल बाघ बचाना था ।

**E5 Output:** जिम कॉर्बेट्त नागरिक उद्यान भारत में 1936 में हैलेय नागरिक उद्यान के रूप में बुढ़ढा राष्ट्रीय उद्यान ऐन्डेंजरेद बंगाल बाघ रक्षा करते हैं है |

**E6 Output:** जिम कॉर्बेट नशनल पार्क को भारत में प्राचीन राष्ट्रीय पार्क हैं और जेखिम में डाला गया बंगाल शेर रक्षा करने के लिए हैलौ नशनल पार्क के रूप में 1936 में स्थापित किया गया था

**Table 3. MT Systems**

| Engine | Unigrams | Bigrams | Trigrams | Prob. Sum |
|--------|----------|---------|----------|-----------|
| E1 | 26 | 25 | 24 | 0.820383 |
| E2 | 32 | 31 | 30 | 0.824706 |
| E3 | 32 | 31 | 30 | 0.043523 |
| E4 | 31 | 30 | 29 | 0.232321 |
| E5 | 29 | 28 | 27 | 0.256545 |
| E6 | 25 | 24 | 23 | 0.564544 |

Table 3 shows the n-gram statistics of these sentences and also shows the sum of cumulative probabilities of these trigrams. By looking at the data we can rank the system according to their probabilities.

## 4. EVALUATION

To evaluate the performance of our system we collected 1300 sentences from tourism domain. These sentences were not part of 35000 that were used to train the models. To validate our results we compared the ranks of the system with the ranks given to MT systems by a human evaluator. The human evaluator used a subjective human evaluation metric that we used by Joshi et al. [16]. This metric evaluated an MT output on ten parameters. These were:

1. Translation of Gender and Number of the Noun(s).
2. Identification of the Proper Noun(s).
3. Use of Adjectives and Adverbs corresponding to the Nouns and Verbs.
4. Selection of proper words/synonyms (Lexical Choice).
5. Sequence of phrases and clauses in the translation.
6. Use of Punctuation Marks in the translation
7. Translation of tense in the sentence
8. Translation of Voice in the sentence
9. Maintaining the semantics of the source sentence in the translation
10. Fluency of translated text and translator's proficiency

Each MT outputs were adjudged on these 10 parameters. The human evaluator was asked to give a score on a 5-point scale. The scale is shown is table 4. Each sentence's 10 scores were then averaged to get a single score which was then used to rank MT outputs. Joshi et al. [17] have illustrated the entire working and detailed evaluation of this metric.

We evaluated the system generated ranks with human ranks in three different categories. At first we compared the ranks of all the systems, irrespective of their type. In second category we compared the ranks of only web based systems and in third category we compared the ranks of only MT toolkits or system which had very limited corpora to train and tune themselves.

**Table 4. Human Evaluation Scale**

| Score | Description |
|-------|-------------|
| 1 | Ideal |
| 2 | Perfect |
| 3 | Acceptable |
| 4 | Partially Acceptable |
| 5 | Not Acceptable |

**Table 5. Ranking at Combined Category**

| Engine | LM Ranking | Human Ranking |
|--------|-----------|---------------|
| **E1** | **467** | **576** |
| E2 | 290 | 389 |
| E3 | 57 | 75 |
| E4 | 77 | 39 |
| E5 | 186 | 78 |
| E6 | 223 | 143 |

**Table 6. Ranking at Web-Based Category**

| Engine | LM Ranking | Human Ranking |
|--------|-----------|---------------|
| **E1** | **633** | **687** |
| E2 | 432 | 473 |
| E3 | 235 | 140 |

**Table 7. Ranking at MT Toolkits Category**

| Engine | LM Ranking | Human Ranking |
|--------|-----------|---------------|
| E4 | 126 | 265 |
| E5 | 456 | 288 |
| **E6** | **718** | **747** |

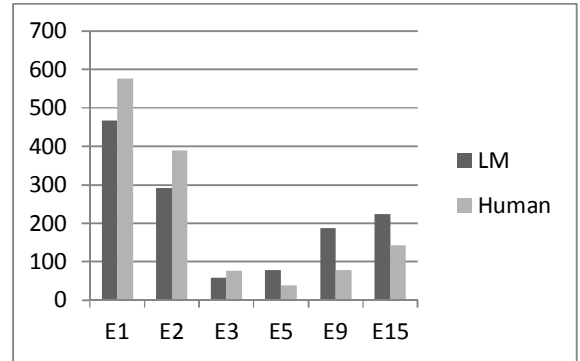

**Figure 2. Ranking at Combined Category**

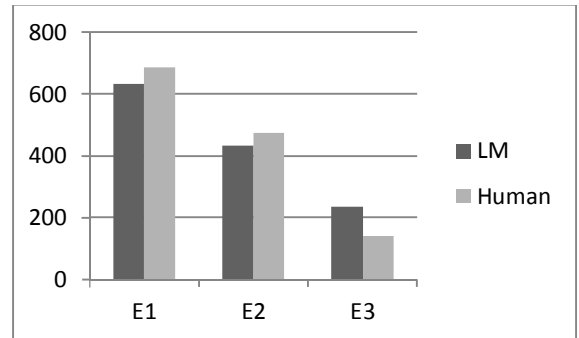

**Figure 3. Ranking at Web-Based Category**

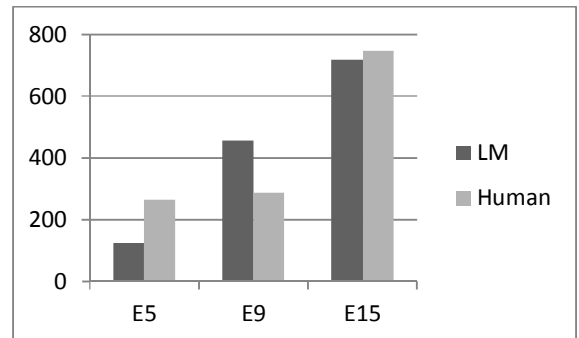

**Figure 4. Ranking at MT Toolkits Category**

In combined category, engine E1 performed better than any other MT engine. It scored the highest rank. Out of 1300 sentences, it managed to score highest rank for 467 sentences. Engine E2 was the second best while engines E3 and E4 did not performed so well. Table 5 shows the results of this ranking. These ranks were similar to the ranks provided by human evaluator.

In web-based category, again E1 and E2 performed better and were the top ranking systems while E3 was the worst. Table 6 shows the results of this study. In MT Toolkits category, E6 performed better than other MT engines and E4 was the worst engine. Table 7 shows the results of this study. Figure 2, 3 and 4 summarizes this data.

## 5. CONCLUSION

In this paper, we have shown the effective use of language models in ranking MT systems. For this we had generated language models for English as well as Hindi. We have also used parallel lexicons to align the trigrams so produced. We also evaluated the MT engines against 1300 sentences which were not part of the training corpus and compared the ranks provided by a human judge. It was found that the ranks produced by LM based ranking and the ranks of human judge were similar. Thus we came to the conclusion that we can use this technique to automatic rank MT systems.

This can be considered as a preliminary study as we still need to perform more experiments to make any sound assumptions. Moreover as an immediate future study we can incorporate part of speech and morphological features into language models and then perform the rank and see if the performance of the system improves or not. Moreover we can also train classifiers and do the ranking. In both these studies this ranking system can be considered as a baseline system.